# Learning Primitive-aware Discriminative Representations for Few-shot Learning

Jianpeng Yang[1],Yuhang Niu[1],Xuemei Xie[1], and Guangming Shi[1]

[1] School of Artificial Intelligence
Xidian University, Xian, China
yangjp@stu.xidian.edu.cn

**Abstract.** Few-shot learning (FSL) aims to learn a classifier that can be easily adapted to recognize novel classes with only a few labeled examples. Some recent work about FSL has yielded promising classification performance, where the image-level feature is used to calculate the similarity among samples for classification. However, the image-level feature ignores abundant fine-grained and structural information of objects that may be transferable and consistent between seen and unseen classes. How can humans easily identify novel classes with several samples? Some study from cognitive science argues that humans can recognize novel categories through primitives. Although base and novel categories are non-overlapping, they can share some primitives in common. Inspired by above research, we propose a Primitive Mining and Reasoning Network (PMRN) to learn primitive-aware representations based on metric-based FSL model. Concretely, we first add Self-supervision Jigsaw task (SSJ) for feature extractor parallelly, guiding the model to encode visual pattern corresponding to object parts into feature channels. To further mine discriminative representations, an Adaptive Channel Grouping (ACG) method is applied to cluster and weight spatially and semantically related visual patterns to generate a group of visual primitives. To further enhance the discriminability and transferability of primitives, we propose a visual primitive Correlation Reasoning Network (CRN) based on graph convolutional network to learn abundant structural information and internal correlation among primitives. Finally, a primitive-level metric is conducted for classification in a meta-task based on episodic training strategy. Extensive experiments show that our method achieves state-of-the-art results on six standard benchmarks.

## 1 Introduction

In recent years, deep learning (DL) has achieved tremendous success in various recognition tasks with abundant labeled data [6 ,7, 8, 9, 10, 11, 72]. To train supervised model efficiently, we need lots of annotated samples that are expensive and time-consuming to obtain. Therefore, how to recognize novel classes with few labeled samples has attracted more attention. To reduce the reliance on human annotation, few-shot learning (FSL) has been proposed and studied widely [1, 2, 3, 4, 5], which aims to learn a classifier that can be rapidly adapted to novel classes given just several labeled images per class.

FSL attempts to transfer useful knowledge learned from base classes with sufficient labeled data to novel classes with only several examples. Humans can rapidly classify an object into one of several novel classes by recognizing the differences between them. Inspired by this ability of humans, a series of few-shot learning methods adopt metric-based algorithm [4, 5], which learns a global image-level representation in an appropriate feature space and directly calculates the distances between the query and support images for classification. Nevertheless, most of these work measure similarity on image-level feature or feature after the pooling operation for classification, which could destroy the structure of objects and ignore local clues.

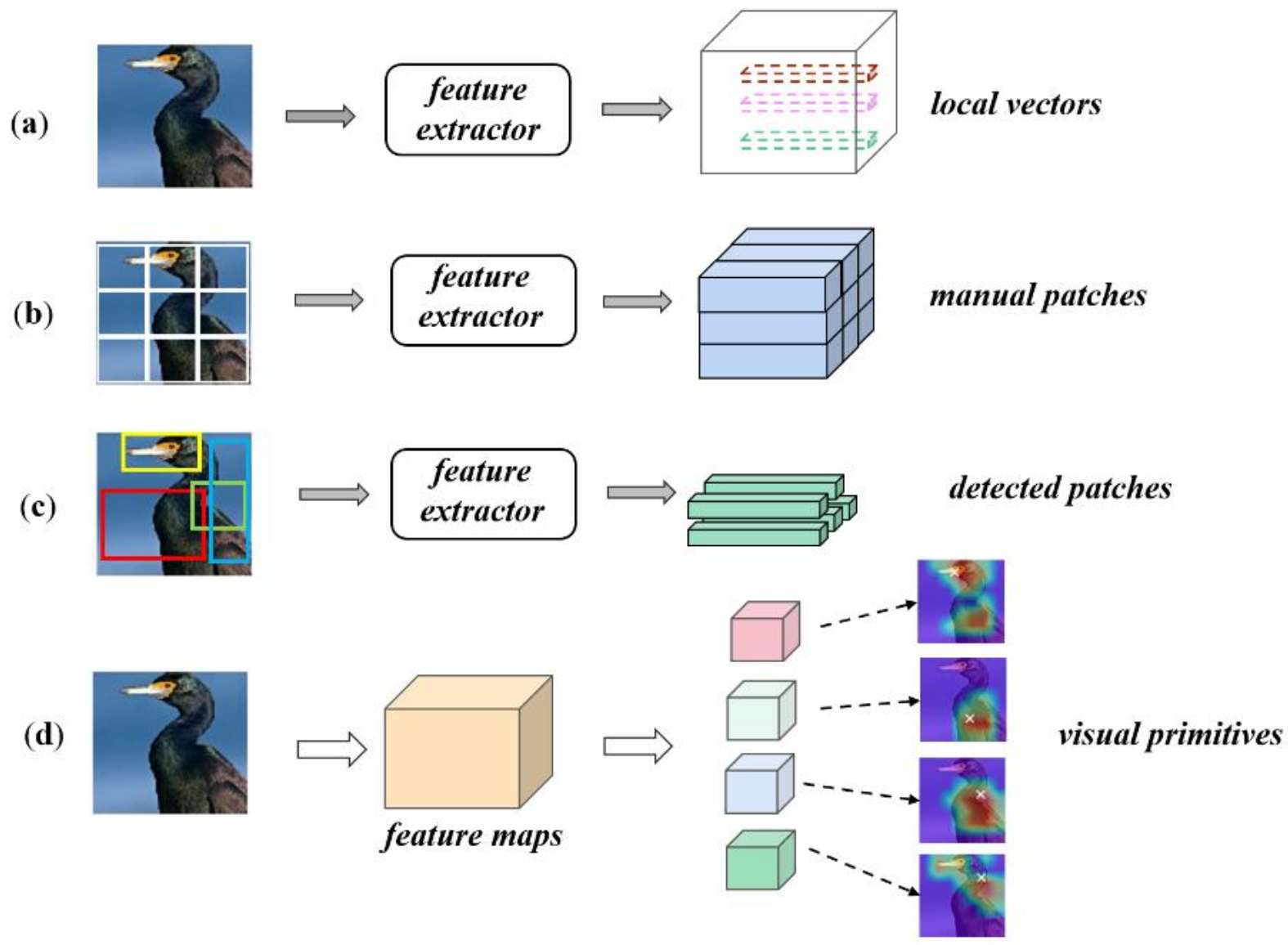

**Fig. 1.** Illustration of our proposed visual primitives and serval types of local representations in few-shot learning. (a) Feature maps are divided into pixel-level local vectors (b) Images are manually divided into a set of patches and these patches are fed into feature extractor to obtain local representations. (c) Bounding boxes of patches are obtained by a detection pretext task. (d) Our proposed visual primitives are composed of visual patterns related to object parts and encode semantic and structural information of objects.

Local representations can usually provide discriminative and transferable information across categories in a few-shot scenario [15]. Due to low-data regime in few-shot learning, tanglesome background and large intra-class variations may make the image-level embedding from the same category far apart in a given metric space [13, 14, 15]. Therefore, local representations (LRs) based methods [16, 15, 14, 13, 17] employ fine-grained feature to measure similarity between query and support images.

Specially, feature maps are manually divided into a set of patches or feature vectors and the measure is conducted on them. However , local patches corresponding to manual grids or pixel-level feature vectors cause too much semantic randomness and introduce redundant information. Also, these methods pay more attention to complex distance functions designed to calculate these dense feature rather than how to capture discriminative and transferable representations.

Revisiting the process that humans recognize new concepts or objects, humans can first learn primitives from plenty of known classes [18] and then apply them to identify novel classes. Specifically, humans fist focus on some similar primitives between known and novel classes, and then mine discriminative clues from novel objects based on these common primitives. Although known and novel classes are non-overlapping, they can share some primitives in common.

In practice, primitives are viewed as object parts, or regions capturing the compositional structure of the examples [19, 23], which doesn't seem to have a clear or fixed boundary. Some researches on interpretability of deep networks show that CNN feature channels often correspond to some visual patterns, such as object parts [20, 21, 22]. Inspired by above studies, CPDE [23] selects single channel of image feature as a primitive by soft composition mechanism. However, this method learn primitives by directly feeding them to general classifier (FC+softmax) , which makes primitive lack of generalization in few-shot scenario. In addition, CPDE only selects top-k channels for classification and cannot take full use of informative clues from all the channels adaptively. TPMN [24] proposes a series of part filters to automatically generate part-aware representations from feature channels and utilizes part-level similarity for classification. Whereas this method has no special trick to encourage feature channels encoding sufficient visual patterns related to object parts, so it is hard to mine discriminative part-aware representations for these part filters. Furthermore, TPMN produces part-aware representations separately so as to neglect internal correlation among them.

Compared to these methods, we thereby compose primitives by selecting feature channels related to object parts rather than single channel, which aggregates visual pattern that is coherent on semantic and spatial relation. Otherwise, it is important to note that all the local representations based methods [16, 15, 14, 13, 17, 24] lose sight of internal correlation among them and structural information of objects. Therefore, we propose to learn and encode internal correlation among visual primitives to improve discriminability of primitive-aware representation.

In this paper, we propose a Primitive Mining and Reasoning Network (PMRN) to learn discriminative and transferable primitive-aware representations for metric-based FSL model based on episodic training mechanism. The main contributions of our work are summarized as follows:

- We develop a Primitive Mining Network (PMN) to learn visual primitives. Firstly, this network guides feature extractor to encode visual patterns related to object parts into feature channels by adding a special Self-supervision Jig-saw（SSJ）task parallelly. Then, it weights and clusters feature channels that are consistent on semantics and spatial location to generate a set of visual primitives through an Adaptive Channel Grouping (ACG) module.

- To capture structural information and inter correlation among primitives, we propose a Correlation Reasoning Network (CRN) to jointly learn the semantic and spatial dependency among primitives by constructing a graph on visual primitives and conduct convolution reasoning operation.
- We design a Task-specific Weight method aiming at measuring the importance of primitives in a metric-based few-shot task, which adaptively generates task-specific weight for weighting primitive-level metric-based classification. To enhance algorithm's generalization across task, all the modules are embedded into meta-task to simulate few-shot scenario by episodic training mechanism.
- Extensive experiments show that PMRN achieves state-of-the-art results on six standard benchmarks.

## 2 Related work

**Few-Shot Learning (FSL)**. Most recent literature about few-shot learning mainly involve two types of research methods, metric-based methods and meta-learning based methods.

The meta-learning based methods [2, 36, 3, 37, 38] optimize a meta-learner utilizing learning-to-learn paradigm [25, 26], which can rapidly adapt to novel classes with just few samples for FSL. An external memory module is employed by [27] to communicate with an LSTM-based meta learner to update weights. Sachin [2] designs an LSTM-based meta-learner as an optimize replacing the SGD optimizer to learn a task-specific initialization for the model. MAML [3] and some variants [31, 29, 30] aim to learn a better parameter initialization that can be quickly adapted to a novel task.

The metric-based methods [32, 4, 33, 12, 5, 34, 35] learn the feature representations for input samples in an appropriate embedding space, where the similarity between images is calculated among different classes through diverse distance metrics. The application of Metric-based method to few-shot learning is first proposed by Koch [32], which aims to generalize representations to novel classes through a Siamese Neural Network. MatchingNet [4] utilizes episodic training strategy and selects cosine similarity as metric to solve FSL problem. ProtoNet [5] regards the mean value of each class's embedding as prototype and calculates Euclidean distance between support and query samples for classification. Our proposed PMRN belongs to metric-based methods, but our method adaptively mine and exploit potential local representations related to object parts, which is more discriminative and transferable among the base and novel categories than the image-level representations utilized by above methods.

**Dense local Feature based FSL.** In contrast to previous methods, some FSL work [13, 14, 16, 15, 17] focus on local representations and try to exploit the discriminative ability of local patches. Specifically, the local patch is considered as each spatial grid in the feature map and all the patch-level distance is aggregated as result. DN4 [13] introduces Naive-Bayes Nearest Neighbor into FSL and computes image-level similarity via a k-nearest neighbor search over local patches. DC [14] predicts for each local features and calculates the average of results as the final prediction. ATL-Net [16] proposes to adaptively select important semantic patches by an episodic attention mechanism. DeepEMD [15] conducts a many-to-many matching method among local patches via the earth mover's distance. MCL [17] consider the mutual affiliations between the query and support features to thoroughly affiliate two disjoint sets of dense local features. Nevertheless, these methods manually preset local patch as each grid or each spatial location in the feature map, which is semantically random and doesn't have any interpretability. Also, dense local patches mean lots of computation in the phase of measure and redundant information. Our method adaptively mine and compose spatially and semantically related visual pattern as primitives in each episode. Therefore, local representations appear in the form of several key primitives related to object parts, which is semantic naturally and leads to better interpretability.

**Interpretability of deep networks.** Some work [22, 20, 21] show that channels of feature extracted by deep networks correspond to some certain visual patterns in the object. Such as object parts, etc. [20]. In this paper, inspired by these studies, we aggregate channels corresponding to certain visual pattern as a primitive, and compute primitive-level similarity between query set and support set.

**Self-supervised learning (SSL).** Manual labels are expensive and time-consuming to collect in practice. SSL aims at learning representations from structural information in the object itself without label [39, 40]. Recently, some work [41, 23, 42] introduce SSL methods such as predicting the rotation and the relative position to FSL. In this paper, we propose to encourage the backbone to learn visual patterns related to object parts by using SSL loss as a regularizer.

## 3 Method

In this section, we first introduce the general settings of few-shot learning method. Then we introduce our proposed Primitive Mining and Reasoning Network (PMRN) concretely, and it is composed of Primitive Mining Network (PMN) in Sec. 3.2 and visual primitive Correlation Reasoning Network (CRN) in Sec. 3.3.

### 3.1 Description and Formulation of few-shot learning.

In few-shot learning scenario, three sets of data are provided, support set $S$, query set $Q$, and an auxiliary set $B$. The support set $S$ contains $N$ novel classes, and each class has $K$ samples. FSL aims at recognizing an unlabeled query sample $q \in Q$ into one of the $N$ novel classes of $S$, and we call such a task as an $N$-way $K$-shot task. However, the support set $S$ only has several labeled samples per class, and an auxiliary set $B$ is employed to train the model thorough episodic training mechanism for learning transferable knowledge. In episodic training, the auxiliary set $B$ is divided into many $N$-way $K$-shot tasks $T$ randomly (also called episodes), where each $T$ contains an auxiliary support set $B_S$ and an auxiliary query set $B_Q$. Each $B_Q$ contains $M$ samples per class. During the training process, the model is forced to conduct hundreds of tasks $T$ to simulate the test scenario, in order to obtain the generalization ability across tasks and learn transferable knowledge that can be used in new $N$-way $K$-shot tasks. Note that auxiliary the set $B$ contains abundant labeled classes and samples, but its label space is disjoint with the set $S$ and $Q$.

### 3.2 Primitive Mining Network

As introduced in Sec. 1, our proposed visual primitive shows its discriminability and transferability in few-shot learning. So, we replace general local representation with visual primitive in metric-based classification and design a Primitive Mining Network (PMN) to learn visual primitive. The detail of architecture is shown in figure 2.

PMN is consist of a Self-supervision Jigsaw (SSJ) task and an Adaptive Channel Grouping (ACG) module. SSJ guides feature extractor to encode visual patterns related to object parts into feature channels. ACG adaptively weights and clusters feature channels that are consistent on semantics and spatial location to generate a set of visual primitives.

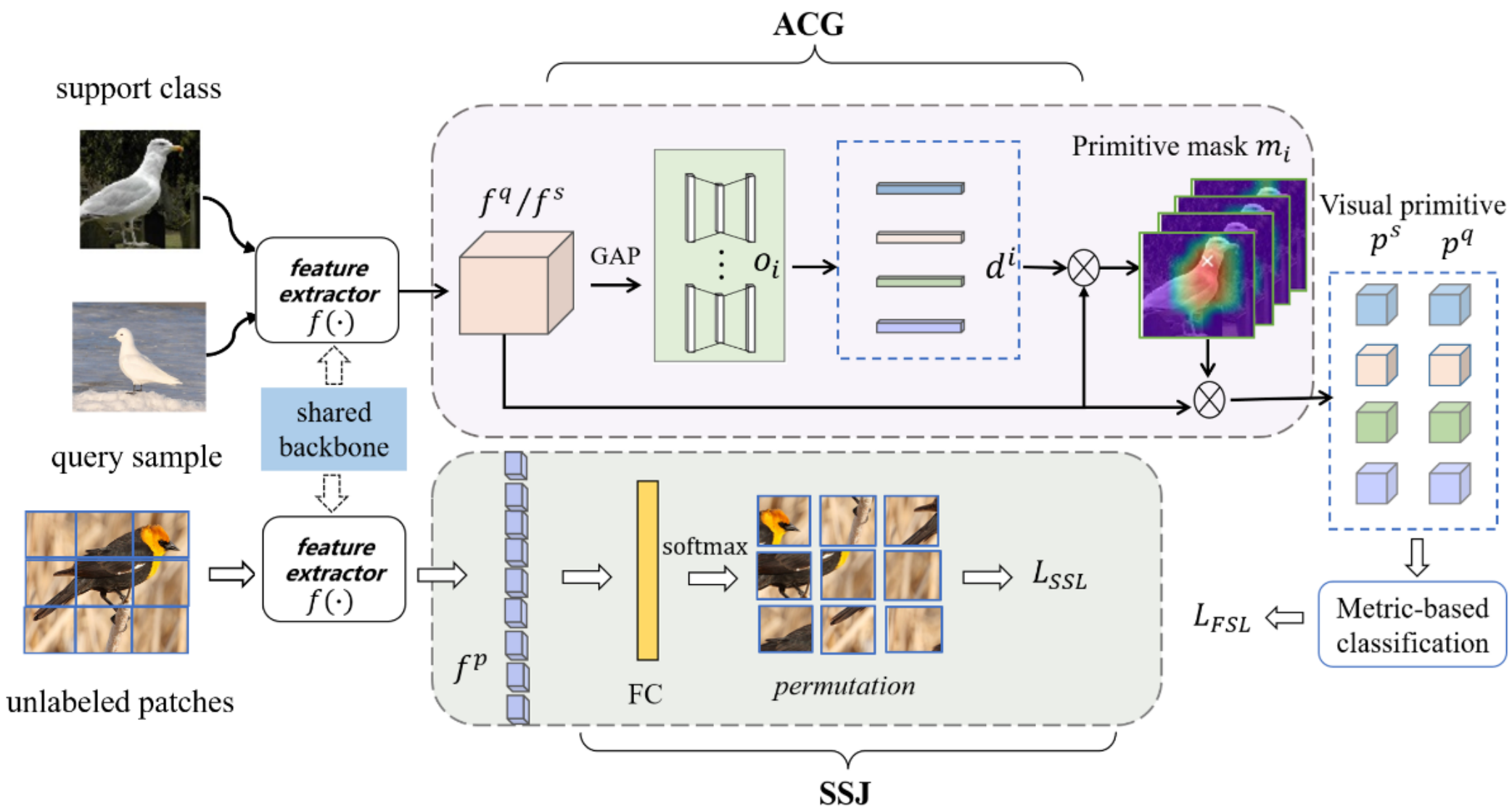

**Fig. 2.** The architecture of Primitive Mining Network (PMN). It is composed of a Self-supervision Jigsaw (SSJ) task and an Adaptive Channel Grouping (ACG) module.

**Self-supervision Jigsaw task.** To assist the primitive mining operation, we adopt a special self-supervision jigsaw task loss [28] to FSL model as a regularizer, which encourage model to recognize related location of image patches by learning visual pattern corresponding to object parts. Note that self-supervision auxiliary task does not need any extra part annotation. SSJ is added in the model parallelly, and its detailed description is shown in figure 2.

Concretely. we first divide an input image $x \in B = \{(x_i, y_i)\}_i^n$ into $h \cdot w$ patches along rows and columns. Afterwards, these patches are permuted randomly as input $x^p$ and we get index of the permutation as the target label $y^p$. The train goal is to predict each index of permutation for patches of an image. Then, all permuted patches are fed into feature extractor $f(\cdot)$ to obtain $h \cdot w$ features and we concatenate the permuted features. Finally, a FC layer for classification with a cross-entropy loss can be utilized to train the model. Let's denote the concatenated features as $f^p$, and the classification loss can be formulated as the negative log-probability:

$$L_{ssl} = -\sum_{x \in B} \log p\,(y^p | f^p) \tag{1}$$

where $f^p \in R^{h \cdot w \cdot d}$, $h \cdot w$ is the number of patches and $d$ is the dimension of feature map. In practice, $h \cdot w$ is set as 3x3 and the number of index reach 9!. To reduce the degree of difficulty, we reduced it to 35 as the procedure proposed in [28], which grouped the possible permutations based on the hamming distance.

**Adaptive Channel Grouping module.** The above process implicitly encode visual pattern related to object parts, which matches well with ACG from the internal mechanism. ACG synthesizes a set of visual primitives from these feature maps adaptively by clustering and weighting a set of spatially and semantically consistent visual patterns encoded in feature channels. The concrete structure of ACG is shown as figure 2 and figure 3.

In each episode, all the images from support and query set $B_S = \{(x_j^s, y_j^s)\}, j \in [1, NK], B_Q = \{(x_j^q, y_j^q)\}, j \in [1, NM]$ are fed into feature extractor $f(\cdot)$ to obtain a collection of features $\Omega = \{f_i^s\}_i^{NK} \cup \{f_i^q\}_i^{NM}$, and the dimension of these feature is $H \times W \times C$, where $H$, $W$, $C$ indicate height, width, and the number of feature channels. Afterwards, a set of parallel channel grouping operations are employed to weight and cluster feature into $k$ groups of feature channels to generate $k$ primitives $P = \{p_1, p_2, \dots p_k\}$ for each sample in $\Omega$, and these operations are denoted as $O = \{o_1, o_2, \dots, o_k\}$.

Intuitively, different feature channels encode different visual patterns, which contributes to different primitives. Therefore, each channel grouping operation $o_i$ is responsible for the generation of a set of weights for each primitive:

$$D_i = o_i(f), i \in [1, k] \tag{2}$$

where $f \in \Omega$ , and $D_i$ is a set of weights $[d_1^i, d_2^i, \cdots d_c^i]$ for the generation of $i^{th}$ primitive $p_i$. After that, we separately employ each set of weights to cluster all the channels into $k$ groups spatially and semantically consistent channels and obtain $k$ primitive masks as follows:

$$m_i = \sigma\left(\sum_{j=1}^{c} d_j^i . a_j\right), i \in [1, k] \quad (3)$$

where $a_j$ is the $j^{th}$ channel map of feature$f \in \Omega$, and the multiplication operation here is conducted between a scalar $d_j^i$and a matrix $a_j$. We use sigmoid function $\sigma(\cdot)$ on the weighted channel maps to generate the $i^{th}$ primitive mask $m_i \in R^{H \cdot W}$, which covers activation region belonging to $i^{th}$ primitive $p_i$. Finally, each feature $f \in R^{H \cdot W \cdot C}$ is filtered through $k$ primitive-level attention masks along channel dimension to acquire the initial primitives:

$$p_i = f \otimes m_i, i \in [1, k] \quad (4)$$

where $\otimes$ denotes the element wise multiplication between $m_i \in R^{H \cdot W}$and each channel of $f \in R^{H \cdot W \cdot C}$. Note the dimension of primitive $p_i \in R^{H \cdot W \cdot C}$ is the same as feature $f \in R^{H \cdot W \cdot C}$.

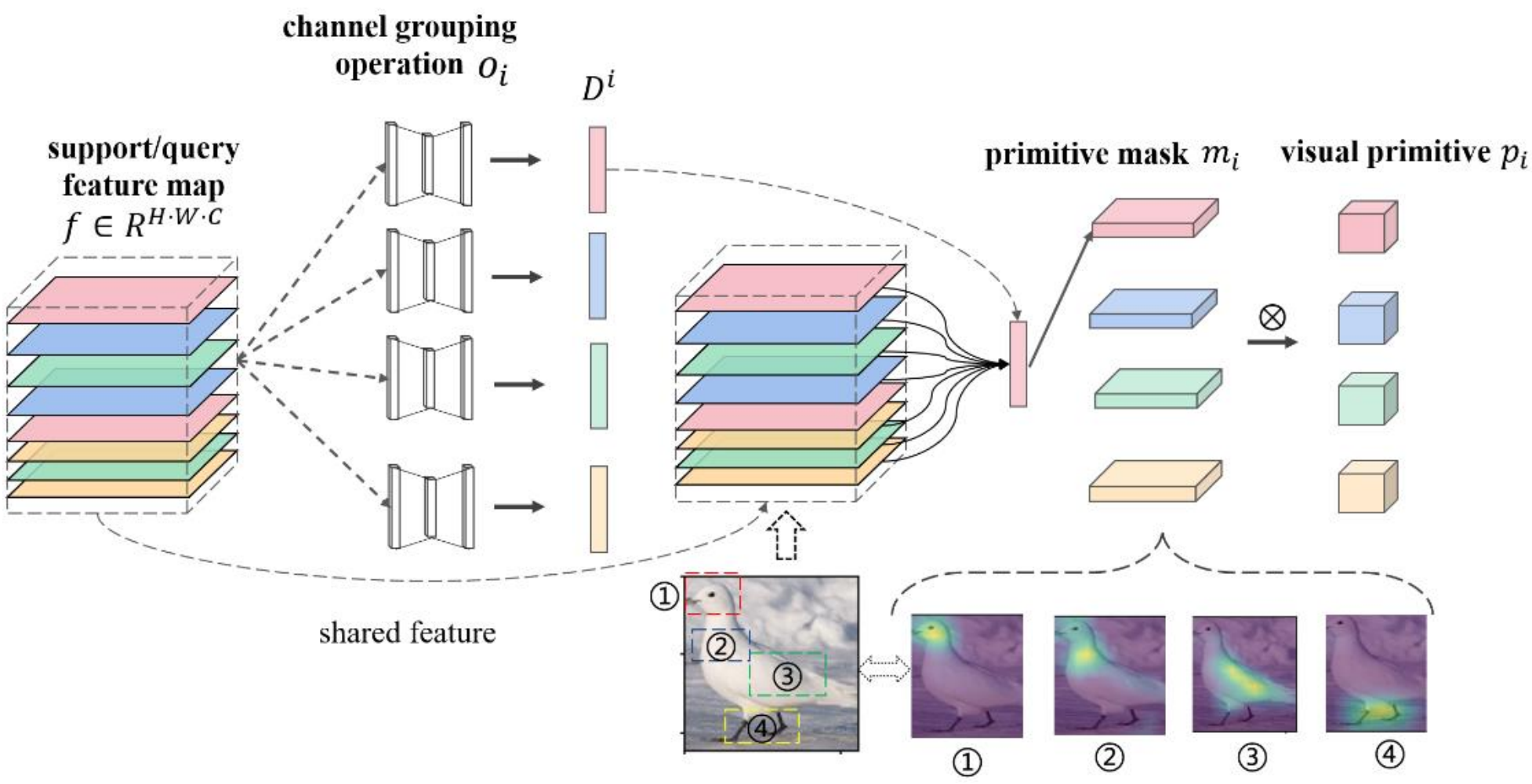

**Fig. 3.** Illustration of Adaptive Channel Grouping module (ACG). We only select a set of weight to show the process of clustering and weighting. $\otimes$ denotes the element wise multiplication between $m_i \in R^{H \cdot W}$and each channel of $f \in R^{H \cdot W \cdot C}$.

After above operations, we capture $k$ visual primitives for each image. Moreover, each channel grouping operation is encouraged to adaptively produce a set of weights, so we utilize global average pooling on feature map and employ cascaded two fully connected (FC) layers to learn a set of primitive-specific weights for each visual primitive. Meanwhile, all the images from support set and query set that belongs to a special task are fed to ACG based on the episodic mechanism, which makes the weights mix all the class information involved in a special task. Hence, our proposed algorithm could be task-specific and more suitable for few-shot learning scenario.

In the training stage, we observe that all the primitive masks tend to cover similar spatial and semantic region, which is usually the most discriminative region. However, the diversity of primitive can further provide rich information and contribute to robust recognition, especially for occlusion cases. To avoid the duplication of learned primitives, we utilize a channel grouping diversity loss inspired by [43, 24], and it is formulated as:

$$L_{div} = \sum_{i=1,i\neq j}^{k} \sum_{j=1}^{k} \delta\left(g(p_i), g(p_j)\right) \tag{5}$$

where $\delta(\cdot)$ denotes the cosine similarity function between primitives from a sample, and $g(\cdot)$ is the global average pooling operation in the spatial dimension.

As is shown in loss formulation, $L_{div}$ will be higher if the $i^{th}$ and $j^{th}$ primitive represent similar spatial or semantic region. Therefore, we can train model to get various primitives by minimizing the $L_{div}$ loss. Here we conduct global average pooling rather than calculate dense pixel-level similarity on primitive $p_i \in R^{H\cdot W\cdot C}$ to improve training and computational efficiency.

### 3.3 Correlation Reasoning Network

As for visual primitive, it is different from the characteristic forms in Euclidean space. Besides spatial distance, visual primitives of object also contain rich structural information and intrinsic semantic correlation. Therefore, it is more suitable to describe visual primitive by irregular graph structure data. Like the graph structure of human skeleton data, visual primitives also have similar graph structure. As shown in Figure 4.

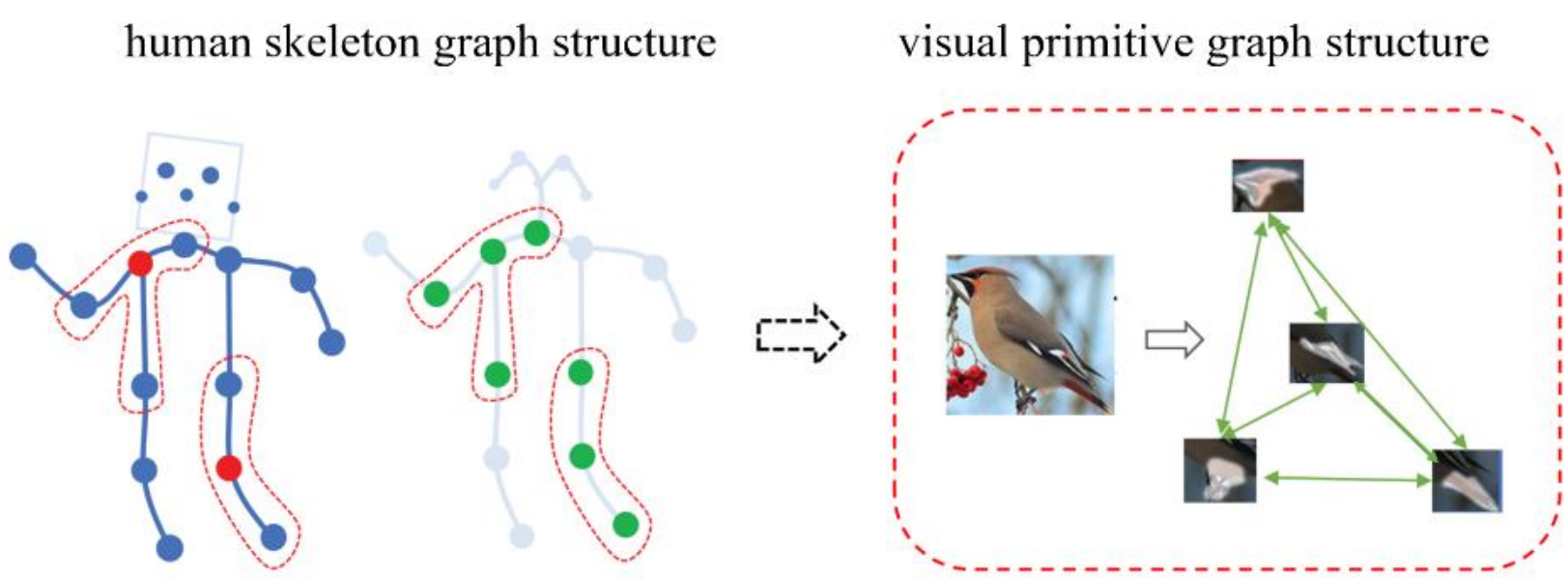

**Fig. 4.** Illustration of visual primitive graph, there are significant spatial and semantic correlations among visual primitives.

To further improve the discriminability and transferability across task of visual primitive, we propose a visual primitive Correlation Reasoning Network (CRN), which constructs a graph structure on visual primitives and designs a special graph convolutional network for reasoning internal correlation among primitives and struc-tural information. In addition, a Task-specific Weight (TSW) method is applied to measure the importance of primitives in primitive-level metric-based classification. The structure of CRN is shown in Figure 5.

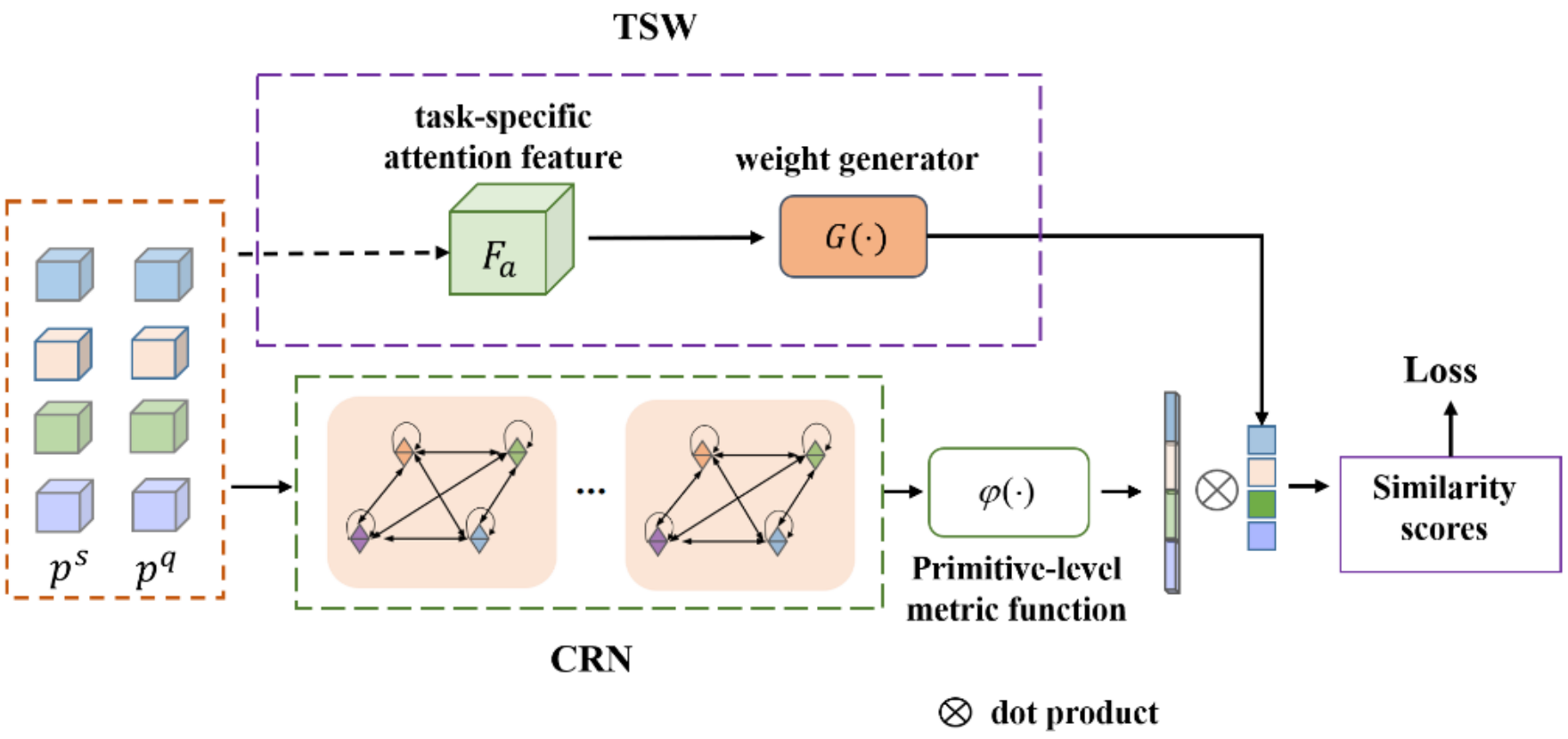

**Fig. 5.** The architecture of visual primitive Correlation Reasoning Network (CRN) and Task-specific Weight (TSW) method.

**Visual primitive graph structure.** Given initial visual primitives $P = \{p_1, p_2, \dots p_k\}$ have been obtained through PMN. we construct a graph on visual primitives at first. The illustration of visual primitive graph structure is shown as Figure 6.

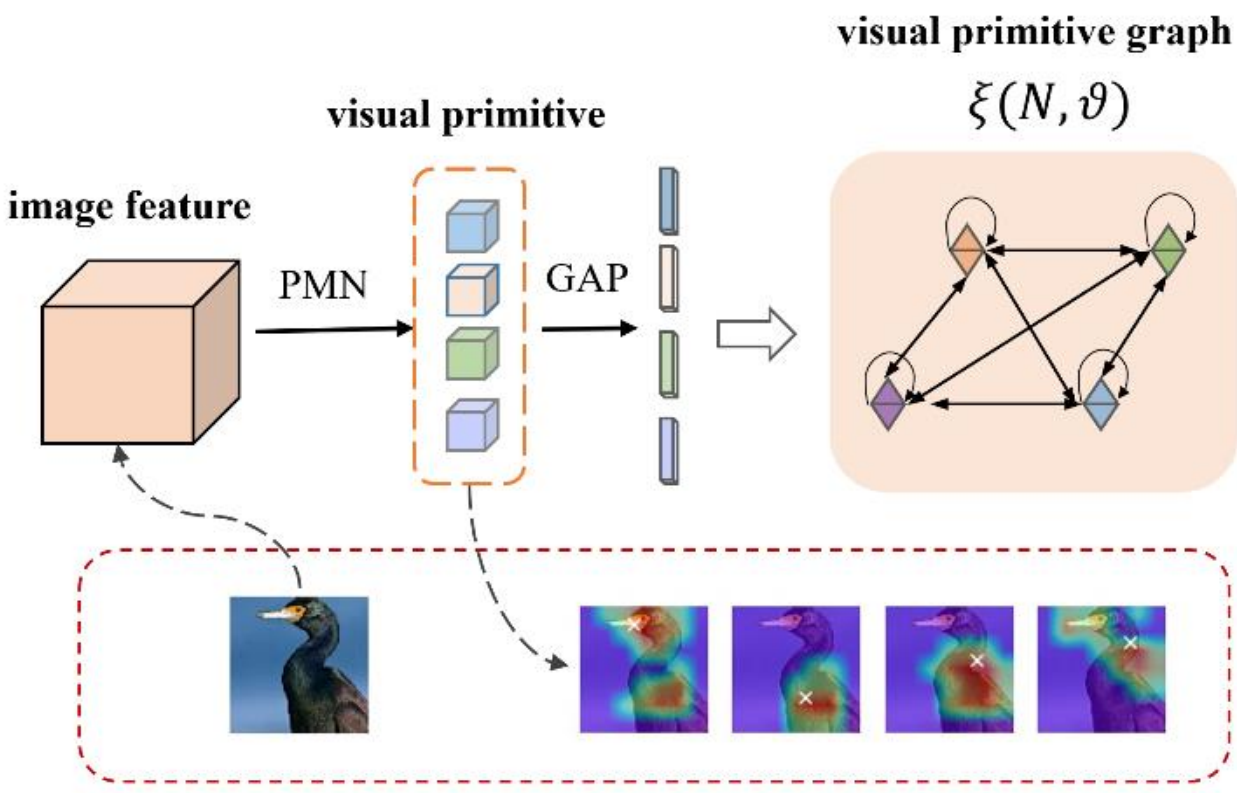

**Fig. 6.** The illustration of visual primitive graph structure. The primitive vectors are embedded into nodes of graph and internal correlation among primitives are regarded as adjacent matrix.

In detail, global average pooling operation is applied to primitives $P = \{p_1, p_2, \dots, p_k\}$ to get $k$ primitive feature vectors as node embedding $N = \{n_1, n_2, \dots, n_k\}, n_i \in R^C$. Then, the similarity matrix of node embeddings should be calculated as adjacent matrix of graph, which encodes semantic and spatial correlation between visual primitives. Moreover, the normalized embedded Gaussian function is adopted to calculate the similarity of the two nodes as follows:

$$S(n_i, n_j) = \frac{e^{\theta(n_i)\omega(n_j)^T}}{\sum_{j=1}^{k} e^{\theta(n_i)\omega(n_j)^T}} \tag{6}$$

where $N$ is the total number of the nodes corresponding to primitives. The dot product is used to measure the similarity of the two nodes in an embedding space. To sum up, visual primitive graph can be formulated as $\xi(N, \vartheta)$, in which $\vartheta$ denotes connection between nodes.

**Adaptive graph convolution layer.** According to the characteristics of visual primitive graph, a special adaptive graph convolutional network is designed to update the visual primitive graph and reason semantic and spatial correlation between visual primitives**.** The specific design of graph convolution layer is shown in Figure 7.

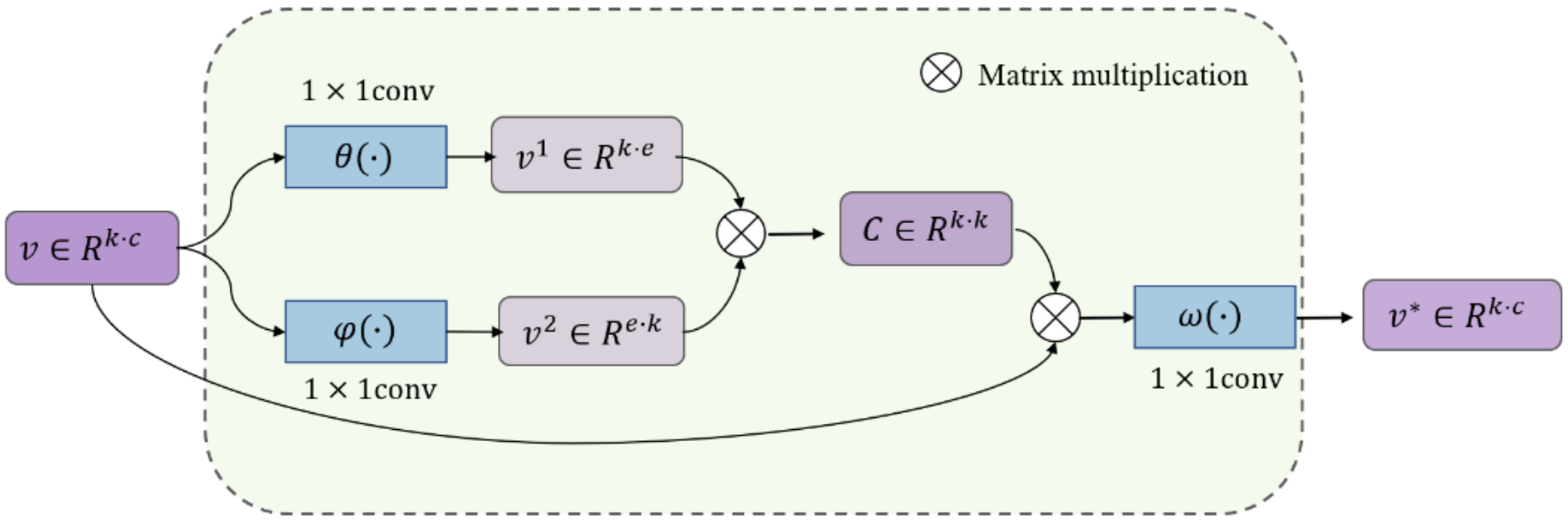

**Fig. 7.** The illustration of **adaptive graph convolution layer** based on visual primitive graph structure.

Specifically, given input vector $v \in R^{k\cdot c}$, $k$ is the number of nodes and $c$ is the dimension of each node embedding, we first project it into $k \times c_e$with two projection functions, i.e., $\theta(\cdot)$and $\varphi(\cdot)$, which are implemented with 1×1 convolutional layer. Then two feature vectors are rearranged and reshaped to an $k \times c_e$ matrix and a $c_e \times k$ matrix separately, which are multiplied to obtain a $K$×$K$ similarity matrix $C$. The element $C_{ij}$ of $C$ represents the similarity of node $n_i$ and $n_j$ , and we conduct softmax operation along the row of matrix for normalization. Based on Eq. (6), the adjacent matrix $C$ can be calculated as follows:

$$C = \sigma((vw_\theta^T) \otimes (w_\varphi v^T)) \tag{7}$$

where $\sigma$ is softmax operation, $\otimes$ is matrix computation, $w_\varphi$ and $w_\theta$ are parameters of $\varphi(\cdot)$ and $\theta(\cdot)$ respectively.

The adaptive graph convolutional layer employs node features $v \in R^{k\cdot c}$ and adjacent matrix $C \in R^{k\cdot k}$as inputs. A single graph convolution layer can be formulated as:

$$v^* = G(v, C) = \varepsilon(C\ v\ w_\omega) \tag{8}$$

where $w_\omega \in R^{c \cdot c^*}$ is the learned weight parameters of 1×1 convolution operation, and $\varepsilon(\cdot)$is ReLU function. Finally, the complete adaptive graph convolutional network based on visual primitives is constructed by stacking multiple graph convolution layers.

**Task-specific weight method**. For each sample in a task *T*, a set of transferable and discriminative visual primitives $P = \{p_1, p_2, \dots, p_k\}$ can be produced through PMN and CRN. Intuitively speaking, the importance of different primitives in a task is different. So that we propose a task-specific weight (TSW) method to measure importance of visual primitives in a task and calculate primitive-level similarity of support and query set. The specific operation of TSW is shown in Figure 8 and Figure 5.

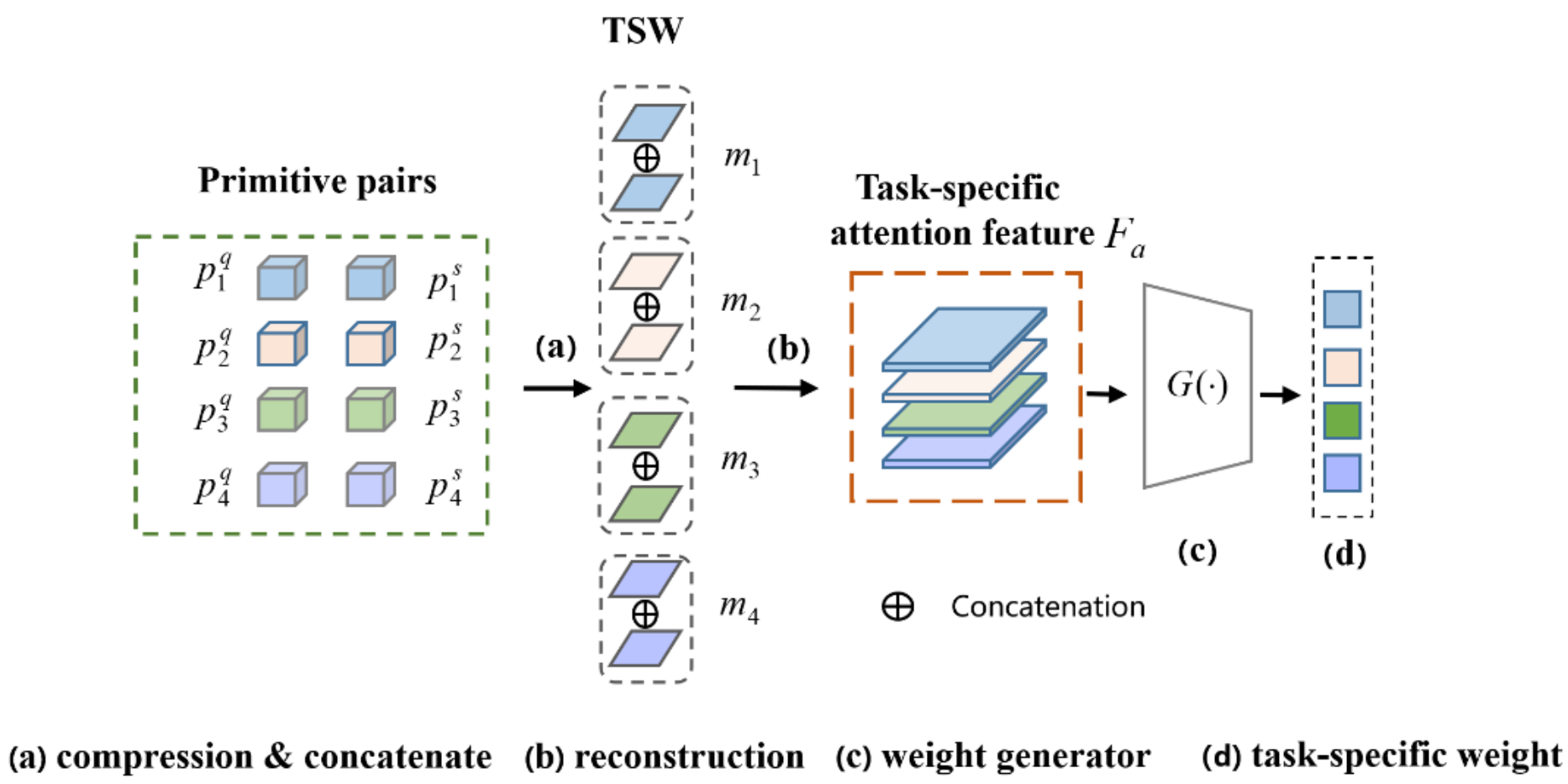

**Fig. 8.** The illustration of task-specific weight (TSW) method. A task-specific attention feature $F_a$ is reconstruct by concatenating primitive maps. The weight generator $G(\cdot)$ adaptively learn task-specific weight $W \in R^k$ for primitive-level metric.

For each query sample $x^q \in B_Q$, we can extract a set of query primitives $P^q = \{p_1^q, p_2^q, \dots, p_k^q\}$, and classify it into one of *N* support classes by calculating similarity between them. For $n^{th}$ support class, we can get *M* set of primitives $P^{s,j} = \{p_1^{s,j}, p_2^{s.j}, \dots, p_k^{s,j}\}_{j=1}^{M}$ from *M* samples and average them to get a primitive-level representation for each class:

$$P^s = \frac{1}{M}\sum_{j=1}^{M} P^{s,j} \tag{9}$$

After that, each support class also have a set of support primitives, which can be denoted as $P^s = \{p_1^s, p_2^s, \ldots, p_k^s\}$. In traditional few-shot model, the final similarity between query sample $x^q \in B_Q$ and $c^{th}$ support class can be calculated simply by summing primitive-level similarity, which can be formulated by:

$$I(x^q, c) = \sum_i^k \varphi\left(g(p_i^q), g(p_i^s)\right) \tag{10}$$

where $\varphi(\cdot)$ is metric function, and it is implemented as cosine similarity in this paper. $g(\cdot)$ is global average pooling operation. The primitive-level similarity is shown in Figure 9.

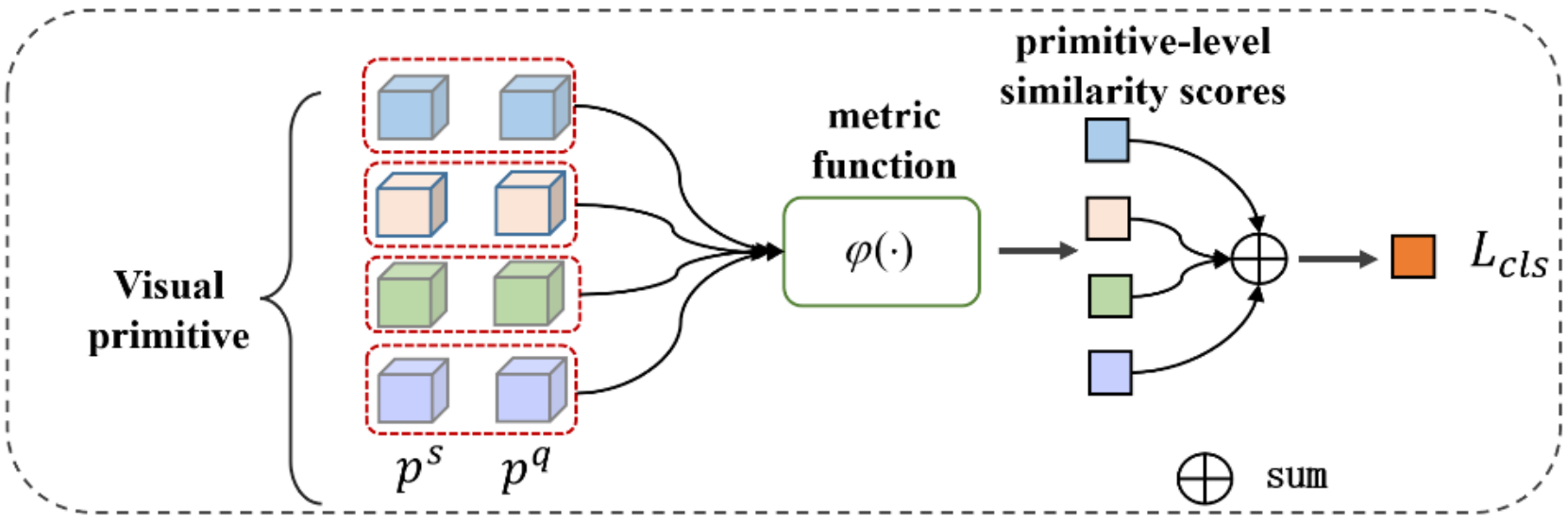

**Fig. 9.** The illustration of primitive-level metric-based classification in few-shot learning task.

As above analysis that contribution of different primitives is discrepant, equally aggregating the primitive-level similarity makes no sense. Therefore, we design a task-specific weight module to adaptively assign appropriate weight for each pair primitive $\{p_i^q, p_i^s\}_{i=1}^k$.Concretely, each primitive in a pair primitive $\{p_i^q, p_i^s\}$ should be compressed into a map to obtain a pair of maps $\{m_i^q, m_i^s\}$. Then we concatenate them along channel dimension as follows:

$$m_i = m_i^q \oplus m_i^s, m_i \in R^{H \cdot W \cdot 2} \tag{11}$$

After that, a task-specific attention feature $F_a \in R^{A \cdot H \cdot W}$ is reconstructed by further concatenating all pairs of maps $\{m_i^q, m_i^s\}_{i=1}^k$ and the detailed process is as Figure 8 shows. The number of channels in the $F_a$ is *A*, and *A=2k*. In contrast to general image-level representation, channels of $F_a$ corresponds to primitive pairs. Thus, we use a weight generator $G(\cdot)$ to measure the importance among different pairs of primitives by learning the importance of different channels of $F_a$. In this way, a set of task-specific weight $W \in R^k$ for each pair of primitives can be generated as follows:

$$W = sigmoid(\gamma(g(F_a), \gamma(m(F_a))) \tag{12}$$

where $g(\cdot)$ and $m(\cdot)$ are global average pooling and max pooling separately. $\gamma(\cdot)$consists of two consistent 1×1 convolution layers and a sigmoid function.

As shown in Figure 5, many task-specific attention feature $F_a$ is applied to train weight generator $G(\cdot)$ across tasks and it adaptively assign higher weight to significant primitives. Therefore, the final similarity should be adjusted to the weighted sum of primitive-level similarity, and Eq. (10) can be changed as follows:

$$I(x^q, c) = \sum_{i=1}^{k} W_i \varphi\left(g(p_i^q), g(p_i^s)\right) \tag{13}$$

### 3.4 Loss and train

Based on episodic training mechanism, all the modules in our proposed Primitive Mining and Reasoning Network (PMRN) are jointly trained end-to-end from scratch without any extra data. Moreover, our framework incorporates self-supervision jigsaw task to few-shot learning by adding an auxiliary self-supervised loss.

For few-shot learning, the final similarity $I(\cdot)$ between query sample $x^q$ and class *c* can be calculated by the weighted sum of primitive-level similarity. Hence, the probability that each query sample $x^q \in Q = \{(x_j^q, y_j^q)\}, j = 1,2, \cdots NM$ could be classified in class *c* can be formulated as:

$$p(y_c|x^q) = \frac{exp(\,I(x^q, c))}{\sum_{c^s=1}^{N} exp(\,I(x^q, c^s))} \tag{14}$$

We compute the classification probability by using a softmax operation and then cross-entropy loss is selected as the few-shot learning loss:

$$L_{cls} = \sum_{x^q \in Q} - log(\,p(y_q|x^q)) \tag{15}$$

consisting of $L_{cls}$, $L_{ssl}$ and $L_{div}$, the loss of our proposed PMRN can be defined as:

$$L = L_{cls} + \lambda L_{div} + \alpha L_{ssl} \tag{16}$$

where $\lambda$ and $\alpha$ are hyper-parameter that controls the importance of the self-supervision loss $L_{ssl}$ and diversity loss respectively.

## 4 Experiments

In this section, we first introduce datasets involved in our experiments and then present some key implementation details. Afterwards, we compare our methods with the state-of-the-art methods on general few-shot learning datasets and fine-grained few-shot learning datasets respectively. Finally, we conduct qualitative analysis and show some ablation experiments to validate each module in our network.

### 4.1 Dataset Description

To evaluate the performance of our proposed PMRN, we conduct expensive experiments on a widely used few-shot learning dataset and five fine-grained datasets:

**miniImageNet** [7] consists of 100 classes with 600 images selected from the ILSVRC-2012 [50]. We follow the split utilized by [2] and take all the classes into 64, 16 and 20 classes as train set, validation set, and test set separately. **Caltech-UCSDBirds-200-2011** [44] contain 11, 788 images from 200 bird classes. Following the splits in [49], we divide them into 100/50/50 classes for train/val/test and each image is first cropped to a human-annotated bound ing box. **Stanford Dogs** [47] contains 120 categories of dogs with a total number of 20, 580 images. **Stanford Cars** [45] contains 196 classes of cars and 16, 185 images. **FGVC aircrafts** [46] has 100 categories about aircrafts and10,000 images are provided. **Oxford flowers** [48] consists of 102 classes about flowers with 8189 images. The detailed illustration of above datasets is denoted as Table 1.

**Table 1.** The splits of all six datasets. N_all is the number of all classes. N_train, N_val and N_novel denotes the number of classes in training set, validation set and test set.

| Dataset | N_all | N_train | N_val | N_novel |
|---|---|---|---|---|
| **miniImageNet** | 100 | 64 | 16 | 20 |
| **Caltech-UCSD Birds** | 200 | 100 | 50 | 50 |
| **Stanford Dogs** | 120 | 60 | 30 | 30 |
| **Stanford Cars** | 196 | 98 | 49 | 49 |
| **FGVC aircrafts** | 100 | 50 | 25 | 25 |
| **Oxford flowers** | 102 | 51 | 26 | 25 |

### 4.2 Implementation Details

PMRN adopts the widely used ResNet-18 [51] as the backbone of our feature extractor$f(\cdot)$, and remove the last pooling layer of it. PMRN is learned by episodic training mechanism and each episode consists of an *N*-way *K*-shot task and 16 query samples are provided for each class. Specifically, there are 5 support images and 80 query images for 5-way 1-shot setting while 25 support images and 80 query images for 5-way 5-shot setting in a single episode. We train the network by ADAM [52] with a learning rate of 0.001. Note that the number of episodes is 100,000 for 5-way and 5-shot setting and 300,000 for 5-way and 1-shot setting. Our model is implemented with the Pytorch [53] based on the codebase for few-shot learning denoted in [41].

In the testing stage, we randomly sample 1000 episodes from the test set and use the top-1 mean accuracy as the evaluation criterion. We report the final mean accuracy with the 95% confidence intervals. All the modules of PMRN are trained from scratch in an end-to-end manner and do not need fine-tuning in the test stage.

In addition, we use data augmentation procedure employed in [52] which achieves a good performance. For self-supervision task, we first randomly crop the original images to get a 255×255 region with random scaling between [0.5, 1.0]. Then we split it into 3×3 regions, which contains nine random patches of size 64×64.The number of primitives $k$ is set as 4 on all the datasets and the default value of hyper-parameter $\lambda$ and $\alpha$ is set as 0.4 and 1.0 separately.

### 4.3 General Few-shot Classification Results

We show experimental result of PMRN for general few-shot learning task on miniImageNet. Table 2 shows the comparison of our proposed PMRN with general few-shot learning methods, including local feature-based methods [16, 33, 35] and the state-of-the-arts.

**Comparison with local feature based methods**. Because our method belongs metric-based few-shot learning branch based on local representations, we first compare our method with some popular metric-based methods that exploits local representations.

The detailed results shown in Table 2 denotes that our method outperforms all these methods, including MCL [17], DN4 [13], ATL-Net [16], DC [14] and DeepEMD [15]. Note that our PMRN achieves amazing margin of 7.56% and 12.69% respectively than the best local-based method MCL [17] in 5-way 5-shot and 5-way 1-shot setting. The key difference is that these local-based methods directly utilize pixel-level or grid-level representations divided from feature maps, where such local patches may contain less informative clues, that is, too much randomness and background noises. In other word, these methods destroy semantic and structural consistency of local features. However, PMRN can adaptively mine and exploit significant visual primitives related to object parts, which makes local representations consistent on semantics and spatial regions. Moreover, PMRN captures internal correlation and structural information among visual primitives by constructing a special graph and graph convolution network, which further enhance the discriminative power of primitives. The significant accuracy gain proves that our primitive-aware representations are more effective than previous local feature-based methods.

**Comparison with the state-of-the-arts.** To We also compare our PMRN with some state-of-the-art methods on miniImageNet.

As Table 2 shows, our proposed PMRN achieves the new state-of-the-art performance on all settings (5-way 5-shot and 5-way 1-shot). Compared with the best method HCT [61], we achieve a remarkable 5.92% performance gain in 5-way 1-shot setting and 2.84% performance gain in 5-way 5-shot setting. The better results indicate improvement of PMRN in general few-shot learning. In contrast to HCT [61] and EASY3-R [62], PMRN systematically design special network and mechanism to learn discriminative visual primitives as local representations for few-shot learning, rather than simply stacks networks or increases the depth of backbone.

**Table 2.** Comparison of our method with the state-of-the-art methods. Few-shot classification (%) results with 95% confidence intervals on miniImageNet.

| Method | Backbone | *mini*ImageNet | |
|---|---|---|---|
| | | 5-way 1-shot | 5-way 5-shot |
| **MatchingNet [4,55,15]** | ResNet-12 | 65.64 | 78.72 |
| **RelationNet [12,17]** | ResNet-12 | 60.97 | 75.32 |
| **ProtoNet [5,17]** | ResNet-12 | 62.67 | 77.88 |
| **CAN [57]** | ResNet-12 | 63.85 | 79.44 |
| **DN4[13]** | ResNet-12 | 65.35 | 81.10 |
| **DeepEMD [15]** | ResNet-12 | 65.91 | 82.41 |
| **ATL-Net [16]** | ConvNet | 54.30 | 73.22 |
| **DSN [59]** | ResNet-12 | 62.64 | 78.83 |
| **FRN [58]** | ResNet-12 | 66.45 | 82.83 |
| **CPDE [23]** | ResNet-18 | 65.55 | 80.66 |
| **TPMN [24]** | ResNet-12 | 67.64 | 83.44 |
| **CTM [60]** | ResNet-18 | 64.12 | 80.51 |
| **MCL [17]** | ResNet-12 | 67.85 | 84.47 |
| **EASY3-R [62]** | 3xResNet-12 | 71.75 | 87.15 |
| **HCT [61]** | 3xTransformers | 74.62 | 89.19 |
| **PMRN（ours）** | ResNet-18 | **80.54** | **92.03** |

### 4.4 Fine-grained Few-shot Classification Results

To further demonstrates the effectiveness of PMRN, we conduct expensive experiments on various fine-grained datasets for fine-grained few-shot learning task.

As shown in Table 3 to 5, PMRN also achieves new state-of-the-art performance. Compared with best methods on Caltech-UCSD Birds-200-2011, Stanford Dogs and Stanford Cars, PMRN has 3.89%, 24.48% and 14.76% accuracy gain under 5-way 1-shot setting and 1.64%, 18.89% and 2.49% accuracy gain under the 5-way 5-shot setting. In addition, we follow the same setting as Su [41] to train some baselines from scratch on Oxford Flowers and FGVC Aircraft. Compared to these baselines, we achieve remarkable performance gain on all setting.

It is obvious that our proposed PMRN achieves larger performance gain on fine-grained datasets than general datasets. For fine-grained few-shot learning task, fine-grained information and clues could be more competitive due to small inter-class differences and large intra-class differences. Image-level representation cannot capture fine-grained information while grid-level or pixel-level local representation brings lots of redundant information and noises. However, visual primitives in PMRN not only encodes local visual patterns related to object parts, but also capture structural information of objects, which can be more effective to distinguish fine-grained classes.

**Table 3.** Comparison of our method with the state-of-the-art few-shot learning methods on fine-grained dataset. The results with 95% confidence intervals on **Caltech-UCSD Birds**.

| Method | Backbone | Caltech-UCSD Birds | |
|---|---|---|---|
| | | 5-way 1-shot | 5-way 5-shot |
| **MatchNet [41, 4]** | ResNet-18 | 73.49 | 84.45 |
| **MatchNet [49, 4, 15]** | ResNet-12 | 71.87 | 85.08 |
| **RelationNet [41, 12]** | ResNet-18 | 68.58 | 84.05 |
| **RelationNet [12, 41]** | ResNet-34 | 66.20 | 82.30 |
| **ProtoNet [41, 5]** | ResNet-18 | 72.99 | 86.64 |
| **MAML [3, 51]** | ResNet-18 | 68.42 | 83.47 |
| **SCA+MAML++ [63]** | DenseNet | 70.33 | 85.47 |
| **Baseline [41]** | ResNet-18 | 65.51 | 82.85 |
| **Baseline++ [41]** | ResNet-18 | 67.02 | 83.58 |
| **S2M2 [64]** | ResNet-18 | 71.43 | 85.55 |
| **DeepEMD [15]** | ResNet-12 | 75.65 | 88.69 |
| **DEML [65]** | ResNet-50 | 67.28 | 83.47 |
| **ATL-Net [16]** | ConvNet | 60.91 | 77.05 |
| **Cosine classifier [41]** | ResNet-18 | 72.22 | 86.41 |
| **DSN [67]** | ResNet-12 | 80.80 | 91.19 |
| **FRN [58]** | ResNet-12 | 83.16 | 92.59 |
| **CPDE [24]** | ResNet-18 | 80.11 | 89.28 |
| **CFA [66]** | ResNet-18 | 73.90 | 86.80 |
| **MCL [17]** | ResNet-12 | 85.63 | 93.18 |
| **PMRN（ours）** | ResNet-18 | **89.25** | **94.82** |

**Table 3.** Comparison of our method with the state-of-the-art few-shot learning methods on fine-grained dataset. The results with 95% confidence intervals on **Stanford Cars** and **Stanford Dogs**.

| Method | Stanford Cars | | Stanford Dogs | |
|---|---|---|---|---|
| | 5-way 1-shot | 5-way 5-shot | 5-way1-shot | 5-way 5-shot |
| **MatchNet [4, 68]** | 34.80 | 44.70 | 35.80 | 47.50 |
| **GNN [69, 68]** | 55.85 | 71.25 | 46.98 | 62.27 |
| **DN4 [13]** | 61.51 | 89.60 | 55.85 | 63.51 |
| **ProtoNet [5, 68]** | 40.90 | 52.93 | 40.90 | 48.19 |
| **MAML [3, 70]** | 47.22 | 61.21 | 44.81 | 58.68 |
| **RelationNet [12, 70]** | 47.67 | 60.59 | 43.33 | 55.23 |
| **MML [71]** | 72.43 | 91.05 | 59.05 | 75.59 |
| **ATL-Net [16]** | 67.95 | 89.16 | 54.49 | 73.20 |
| **CovaMNet [68]** | 56.65 | 71.33 | 49.10 | 63.04 |
| **PABN+cpt [69]** | 54.44 | 67.36 | 45.65 | 61.24 |
| **LRPABN+cpt [69]** | 60.28 | 73.29 | 45.72 | 60.94 |
| **MATANets [71]** | 73.15 | 91.89 | 55.63 | 70.29 |
| **PMRN (ours)** | **87.91** | **94.38** | **80.11** | **89.18** |

**Table 4.** Comparison of our method with some baselines methods on fine-grained dataset. The results with 95% confidence intervals on **Oxford Flowers and FGVC Aircraft**. We re-implement above five methods with ResNet-18 in these two datasets for fair comparison.

| Method | Oxford Flowers | | FGVC Aircraft | |
|---|---|---|---|---|
| | 5-way 1-shot | 5-way 5-shot | 5-way 1-shot | 5-way 5-shot |
| **MatchNet** | 77.53 | 89.03 | 81.74 | 88.48 |
| **ProtoNet** | 76.85 | 89.20 | 79.29 | 91.40 |
| **MAML** | 61.51 | 79.12 | 75.34 | 88.92 |
| **RelationNet** | 69.55 | 85.73 | 76.52 | 87.41 |
| **Softmax** | 78.36 | 91.10 | 74.26 | 89.20 |
| **PMRN (ours)** | **82.43** | **95.75** | **80.65** | **95.33** |

### 4.5 Qualitative analysis

To further evaluate the effectiveness of PMRN and illustrate its mechanism, we conduct special and ample qualitative analysis.

**Local representations related to structural clues**. As shown in Figure 10, the activation regions of visual primitive roughly correspond to certain structural parts or semantic regions of object, such as legs and abdomen of birds.

For the same class, such as support sample and query sample (a), the activation regions of visual primitive are consistent in object structure to a certain extent, which shows that the same visual primitives tend to cover the same structural parts or semantic regions. The above analysis illustrates that PMN can adaptively mine and generate visual primitives corresponding to parts or structure of object, which could be more discriminative local representations, and it possesses certain interpretability and transferability on semantic or spatial location.

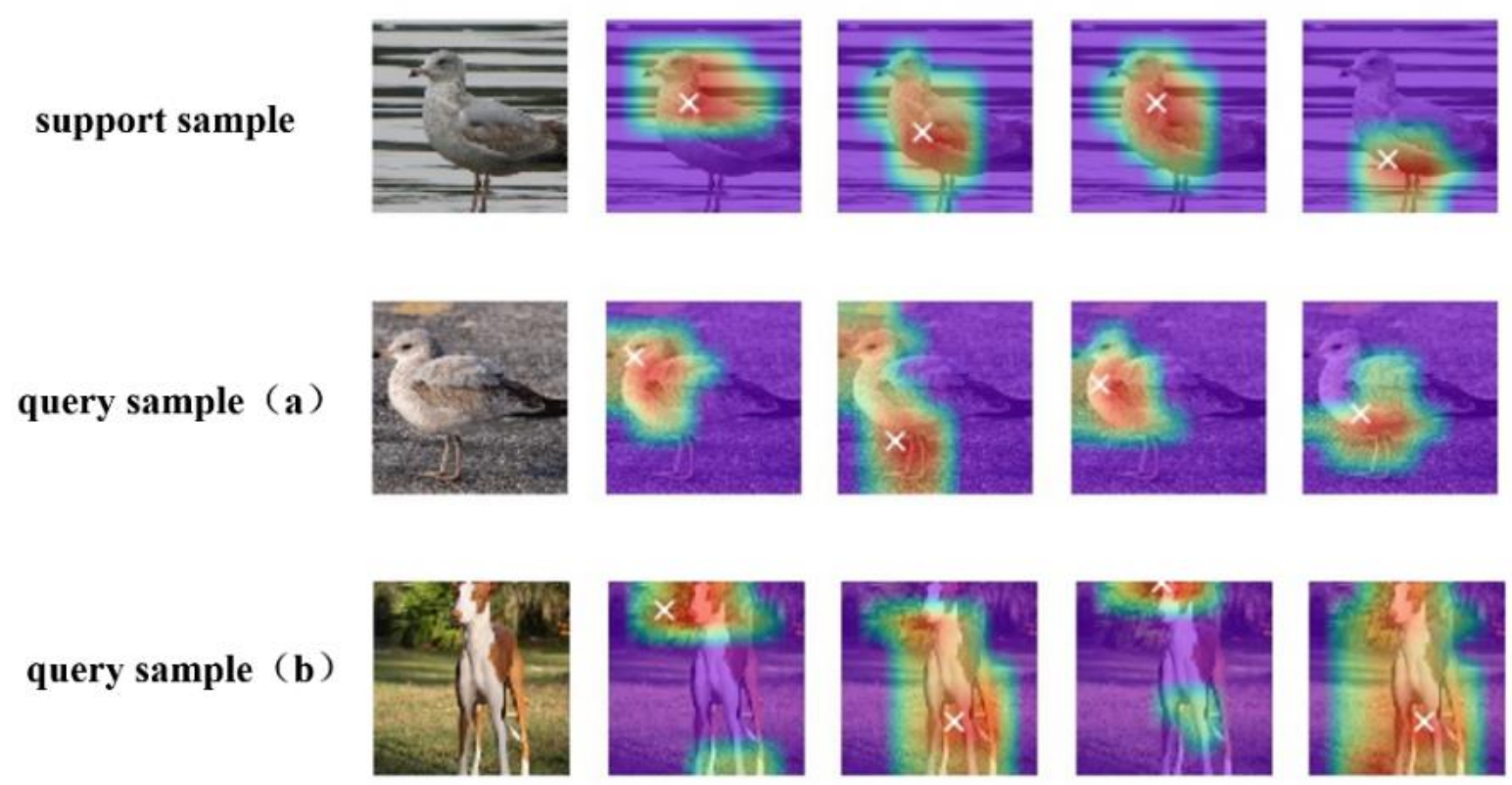

**Fig. 10.** The visual primitive visualization of a support sample and two query samples, where query sample (a) and support sample are of the same class. Note that visual primitives are produced by PMN.

For different classes, such as query sample (b) and support sample in Figure 10, although they have huge differences in structure, visual primitives still roughly cover similar structural parts or semantic regions, for example, visual primitive activating neck and head of a bird also covers the neck and head of a dog. Above analysis reflects the transferability and generalization across tasks of visual primitives. It also shows that visual primitives based on episodic training mechanism can adapt to few-shot scenario well.

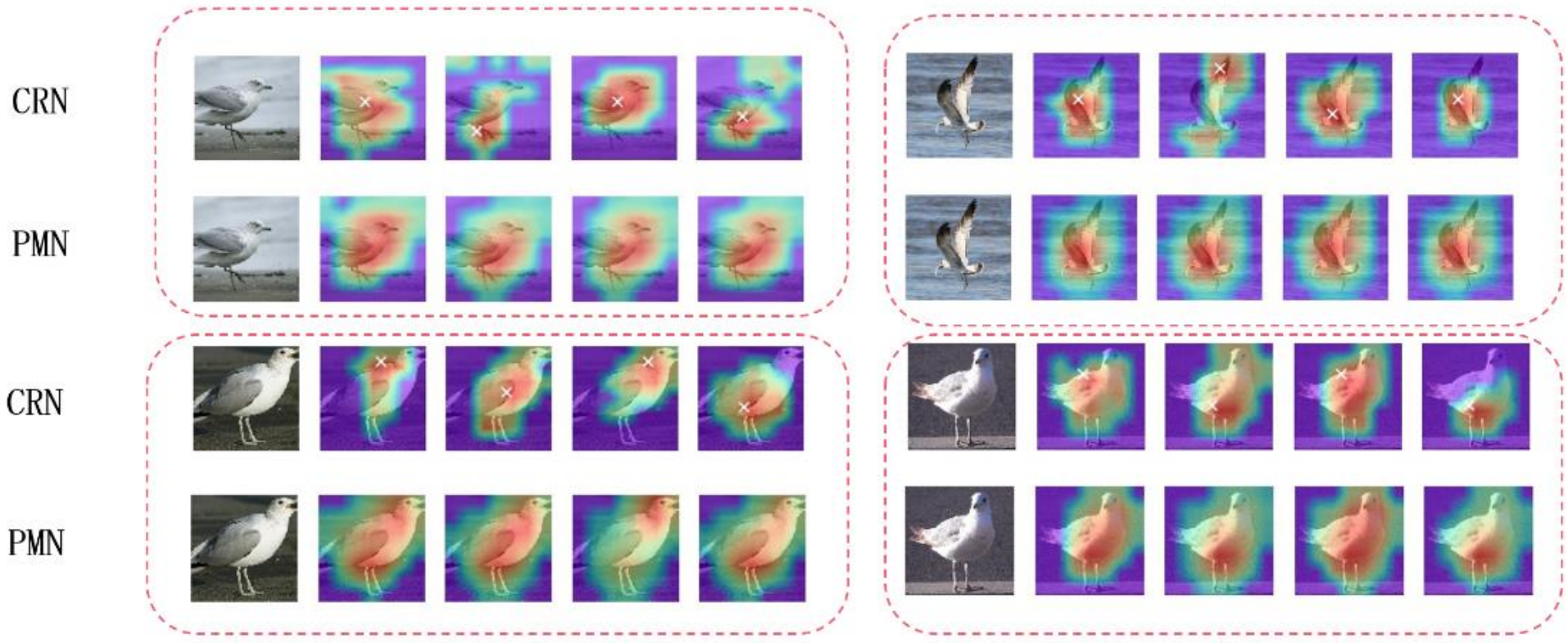

**Fig. 11.** The visualization of visual primitive produced by PMN (second row) and CRN (first row) respectively.

**Inter correlation among visual primitives.** As shown in Figure 11, for several visual primitives with poor interpretability and discriminability produced by PMN, CRN makes them more natural in structural parts and distinguishable in spatial regions. Because visual primitives are related to object parts, strong inter correlation on structure and semantics should be mined necessarily. It can be concluded that spatial and structural information among visual primitives is constrained and optimized by constructing visual primitive graph and reasoning by graph convolution network, so that some confusing visual primitives can be optimized.

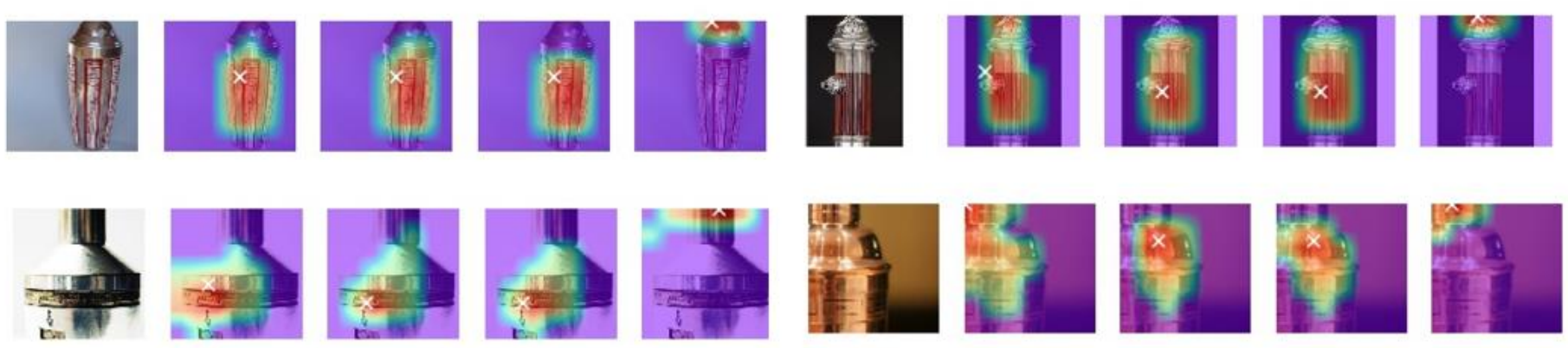

**Fig. 12.** The visual primitive visualization of objects with sample structure.

**The number of visual primitives**. Some objects with simple structure may contain fewer discriminative visual primitives, as shown in Figure 12. The structure of bottle may only contain the upper and lower parts, which will lead to the homogeneity of visual primitive, that is, different visual primitives will cover similar regions or parts. Because PMRN sets a fixed number of visual primitives, such a phenomenon may occur if the number of discriminative visual primitives of object is fewer than our setting. Although a channel grouping diversity loss is applied to avoid this issue, forcing a non-discriminative visual primitive to be generated has a penalty to accuracy. In this case, it is reasonable to generate homogeneous visual primitives for objects with simple structure.

This phenomenon is even more prevalent in general datasets, which contain multiple classes with widely varying structural information, ranging from simple bottles to complex human bodies. To explore the influence of the number of visual primitives on classification accuracy, we conduct simple experiments on Stanford Dogs and FGVC Aircraft datasets. As shown in Figure 13. Our target is to observe the change in few-shot classification accuracy when the number of visual primitives increases within a certain range.

On Stanford Dogs dataset, when the number of primitives rises to 4, classification accuracy reaches the highest value, and the change of classification accuracy tends to be stable, which indicates the number of discriminative visual primitives may be 4. On the FGVC Aircraft dataset, when the number of primitives reaches 3, the change of classification accuracy tends to be stable, so it is inferred that the number of discriminative primitives for aircraft may be 3.

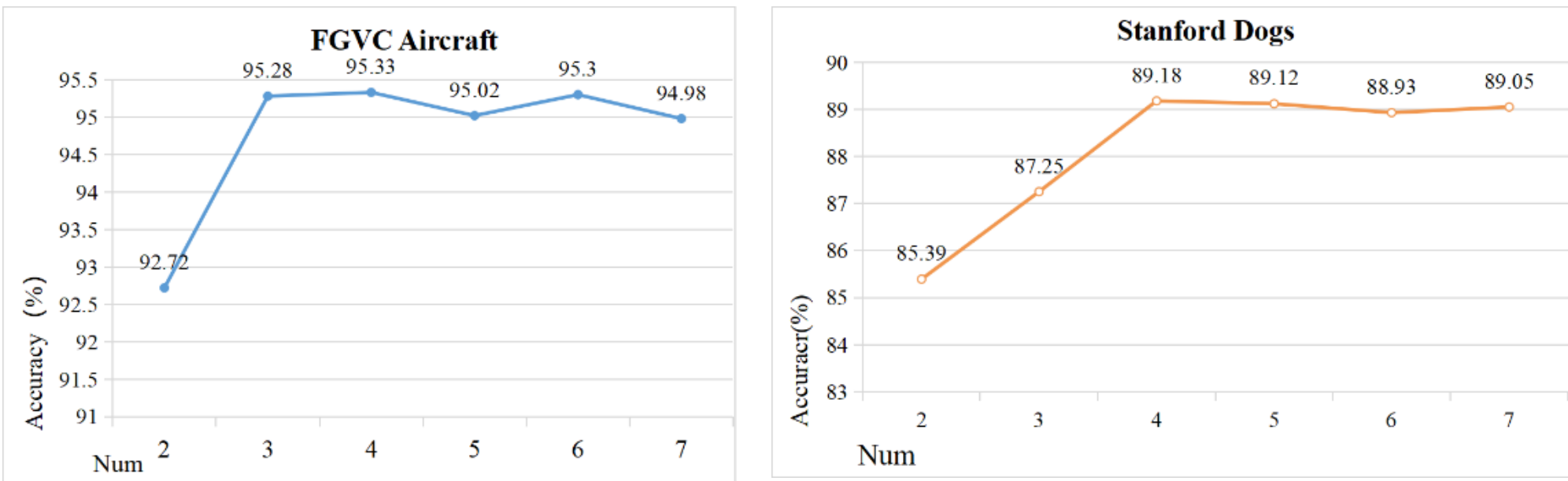

**Fig. 13.** The effect of variation in the number of visual primitives on few-shot learning accuracy on the Stanford Dogs dataset and FGVC Aircraft dataset (5-way 5-shot).

According to above qualitative analysis and quantitative analysis, the number of discriminative visual primitives for different objects is different, which is closely related to object's structure. It is necessary to set enough visual primitives, but computation cost and the complexity of network structure must be considered. However, increasing the number of visual primitives will increase the complexity and calculation cost of network structure. In this paper, the hyperparameter of the number of visual primitives is set to 4, which is a reasonable trade-off. In the future, it is necessary to further explore the a flexible generation mechanism of visual primitives to adapt to real scene. Figure 13 shows more visualization results for other categories and samples.

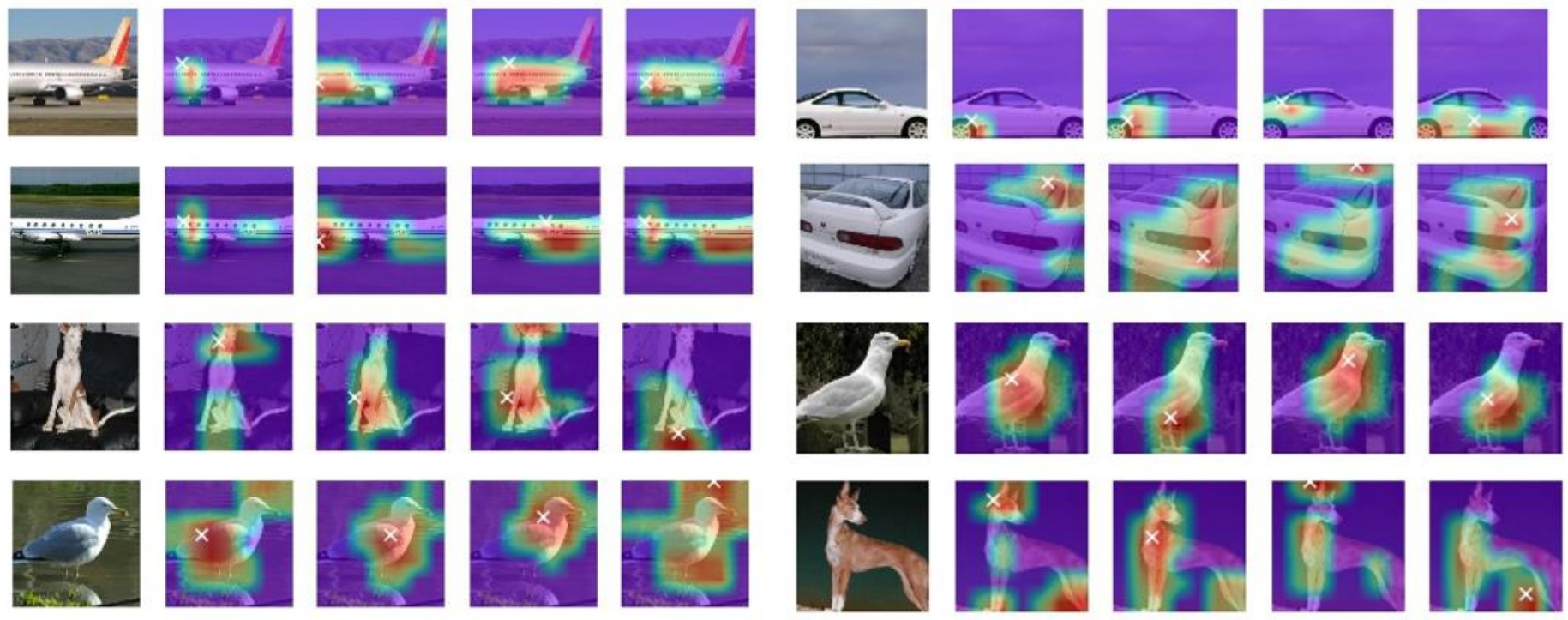

**Fig. 13.** The visual primitive visualization of more classes and samples.

### 4.6 Ablation Study

To assess the effectiveness of each module in PMRN, we conduct detailed ablation studies on the miniImageNet dataset.

We first introduce our baseline as basis reference for validation study of other modules. Specifically, we use the ResNet-18 as the backbone, then the episodic training mechanism is used to classify each query sample into one of the *N* support classes, called N-way K-shot task as the ProtoNet [5] shows. The main difference is that we change similarity function from euclidean distance to cosine similarity. As shown in Table 7, we add various modules on the baseline respectively to verify the effectiveness of each module.

**Table 7.** Ablation results on **miniImageNet** in 5-way 1-shot and 5-way 5-shot settings for the proposed PMRN.

| Method | SSJ | ACG | CRN | TSW | 5-way 1-shot | 5-way 5-shot |
|---|---|---|---|---|---|---|
| baseline | | | | | 61.51 | 75.32 |
| + SSJ | √ | | | | 62.04 | 76.12 |
| + ACG | | √ | | | 64.02 | 79.38 |
| + ACG+CRN | | √ | √ | | 65.57 | 82.25 |
| + ACG+TSW | | √ | | √ | 71.54 | 85.21 |
| + ACG+CRN+TSW | | √ | √ | √ | 77.13 | 89.24 |
| + SSJ +ACG | √ | √ | | | 79.26 | 90.02 |
| + SSJ +ACG +CRN +TSW | √ | √ | √ | √ | 80.54 | 92.03 |

Compared to baseline method, our proposed ACG module improve the accuracy by 2.51% in 1-shot setting and 4.06% in 5-shot setting. This large improvement proves that primitive-level representations possess better discriminative power. Based on primitive-level representations produced by ACG module, we utilize CRN and task-specific weight module (TSW) separately to enhance the model: With the application of CRN module, further improvements by 2.87% in 5-shot setting indicates that considering the internal correlation among primitives is necessary. Meanwhile, TSW achieves a large accuracy gain of 5.25% and 7.52% in 5-shot and 1-shot setting respectively on the top of ACG module, which reveals our proposed task-specific weight generation mechanism can select both transferable and discriminative primitives across tasks indeed. It is worth noting that the combination of CRN and TSW acquire remarkable performance improvement than any single addition of them. The 9.86% and 13.09% performance gain in 5-shot and 1-shot setting can adequately demonstrate they can mutually be optimized and enhanced in the training stage.

In addition, PMN is the combination of SSJ and ACG and acquire remarkable performance improvement than any single addition of them, which indicates they are mutually promoted and optimized. Furthermore, the combination of SJP and ACG achieves better accuracy improvement than the combination of ACG and other modules. It is concluded that the application of SSJ is by no means a simple stack of modules. SSJ guides the feature extractor to mine visual patterns corresponding to object parts, which not only encourage feature to learn more local information, but also improve the generation of visual primitives from underlying learning mechanism. In other words, SSJ encodes visual patterns corresponding to object parts into feature channels during training, and this mechanism perfectly matches the generation of visual primitives based on ACG.

### 4.7 Cross-domain Experiments

Cross-domain test is a challenging task for few-shot learning thanks to the large domain gap between different datasets, which can evaluate the model's ability of transferring knowledge and generalization.

**Table 8.** Cross-Domain experiments results from *mini*ImageNet to CUB.

| Method | 5-way1-shot | 5-way 5-shot |
|---|---|---|
| **cosmax** [7] | 43.06 | 64.38 |
| **ProtoNet** [5] | 47.51 | 67.96 |
| **centroid** [54] | 46.85 | 70.37 |
| **FEAT** [55] | 50.67 | 71.08 |
| **MatchingNet**[4] | 51.65 | 69.14 |
| **Consine classifier** [7] | 44.17 | 69.01 |
| **TPMN** [24] | 52.83 | 72.69 |
| **PMRN** | **58.95** | **72.89** |

We employ the setting in ProtoNet [5] and conduct a cross-domain experiment where our model is trained on miniImagenet and evaluated on the CUB dataset. The detailed results are shown in the Table 8. As the results show, our PMRN improves these baseline models observably under all setting, especially in 1-shot setting, e.g.

6.12% accuracy gain. This demonstrates that our proposed model possess better generalization across domain and the primitive-level local representations are more transferable than image-level representations. Furthermore, capturing the internal semantic correlation among visual primitives and task-specific weight generation mechanism make the model possess remarkable discriminative ability for recognition across task. Hence, our model can be rapidly applied to the novel classes with domain gap.

## 5 Conclusion

Inspired by rapid recognition ability of humans, we research visual primitive as local representations for metric-based few-shot learning. In this paper, we propose a Primitive Mining and Reasoning Network (PMRN) for few-shot learning, and it is composed of Primitive Ming Network（PMN) and visual primitive Correlation Reasoning Network (CRN). PMN adaptively generates primitive-aware representations by mining and clustering visual patterns related object parts based on episodic training strategy. Then visual primitives are applied to conduct primitive-level metric for classification. CRN constructs special visual primitive graph structure and graph convolution network to reason internal correlation among visual primitives. Based qualitative and quantitative analysis, visual primitives in PMRN show remarkable transferable and discriminative power across tasks and domains. Extensive experiments indicate the effectiveness of our method.